\documentclass[conference]{IEEEtran}
\IEEEoverridecommandlockouts
\IEEEoverridecommandlockouts                              

\usepackage{graphicx} 
\usepackage{epsfig} 
\usepackage{mathptmx} 
\usepackage{times} 
\usepackage{amsmath} 
\usepackage{amssymb}  
\usepackage{gensymb}
\usepackage{booktabs} 
\usepackage{multirow}
\usepackage{xcolor}
\usepackage{import}
\usepackage{subcaption}
\usepackage{svg}
\usepackage{tikz}
\usepackage[letterpaper, left=0.639in, right=0.639in, bottom=1.05in, top=0.75in]{geometry}
\usepackage{eso-pic} 
\def\BibTeX{{\rm B\kern-.05em{\sc i\kern-.025em b}\kern-.08em
    T\kern-.1667em\lower.7ex\hbox{E}\kern-.125emX}}
\newcommand{\etal}{\textit{et al.}}
\title{\LARGE \bf
CaLiV: LiDAR-to-Vehicle Calibration of Arbitrary Sensor Setups}

\author{Ilir Tahiraj$^{1,*}$, Markus Edinger$^{2}$, Dominik Kulmer$^{1}$, Markus Lienkamp$^{1}$
\thanks{$^{*}$ Corresponding author {\tt\small ilir.tahiraj@tum.de}}
\thanks{$^{1}$Authors are with the TUM School of Engineering and Design, Chair of Automotive Technology,
       Technical University of Munich.}%
\thanks{$^{2}$with the TUM School of Computation, Information and Technology, Technical University of Munich.}
}

\begin{document}

\thispagestyle{empty}
\pagestyle{empty}

\maketitle

\begin{abstract}
In autonomous systems, sensor calibration is essential for safe and efficient navigation in dynamic environments. Accurate calibration is a prerequisite for reliable perception and planning tasks such as object detection and obstacle avoidance. Many existing LiDAR calibration methods require overlapping fields of view, while others use external sensing devices or postulate a feature-rich environment. In addition, Sensor-to-Vehicle calibration is not supported by the vast majority of calibration algorithms. In this work, we propose a novel target-based technique for extrinsic Sensor-to-Sensor and Sensor-to-Vehicle calibration of multi-LiDAR systems called CaLiV. This algorithm works for non-overlapping fields of view and does not require any external sensing devices. First, we apply motion to produce field of view overlaps and utilize a simple Unscented Kalman Filter to obtain vehicle poses. Then, we use the Gaussian mixture model-based registration framework GMMCalib to align the point clouds in a common calibration frame. Finally, we reduce the task of recovering the sensor extrinsics to a minimization problem. We show that both translational and rotational Sensor-to-Sensor errors can be solved accurately by our method. In addition, all Sensor-to-Vehicle rotation angles can also be calibrated with high accuracy. We validate the simulation results in real-world experiments. The code is open source and available on https://github.com/TUMFTM/CaLiV.

\end{abstract}

\section{INTRODUCTION}
Fusion of sensor data is essential for autonomous robots to interpret their environment. Multi-sensor fusion relies on extrinsic sensor calibration to accurately align sensor measurements. Inaccurate sensor calibration can propagate through the autonomous system, affecting not only fusion tasks such as object detection, but also safety-critical downstream processes such as planning and control~\cite{fusion}. 

The performance of sensor fusion tasks relies primarily on accurate Sensor-to-Sensor (S2S) calibration. However, the global position accuracy of the fused data is highly dependent on Sensor-to-Vehicle (S2V) calibration, making S2V calibration particularly important for downstream tasks such as planning, where accurate object and obstacle positions are required. S2V involves estimating calibration parameters relative to a vehicle or base frame, rather than between individual sensors. In the current state of the art, sensor calibration is primarily addressed in the S2S context. S2V calibration approaches have not received much attention from the research community, even though it has a major impact on the overall system performance and safety. 

Unlike conventional S2S approaches, there are no directly comparable measurements available to find correspondences between a sensor frame and a vehicle frame to perform a calibration. Therefore, almost all S2V calibration frameworks use motion-based approaches such as Simultaneous Localization and Mapping (SLAM) or hand-eye calibration to solve the S2V problem. This is primarily because the motion itself serves as the correspondence to be identified between the sensor and the vehicle.

A similar problem occurs in sensor calibration when dealing with non-overlapping sensor setups. The calibration becomes significantly more challenging with sensors that do not share a common field of view (FoV) and, therefore, never simultaneously observe the same features. This is common in applications involving large vehicles, such as buses or trains. Accordingly, non-overlapping sensor calibration frameworks also use motion-based methods, including SLAM or hand-eye calibration~\cite{horaud}. 

SLAM-based calibration methods generate a map of the environment through point cloud matching and implicitly solve for the extrinsic parameters. These approaches rely heavily on a feature-rich environment, which may not always be available. Hand-eye calibration methods are less dependent on specific environmental features, but do not achieve the calibration accuracy of SLAM algorithms~\cite{Das2022}. In addition, these motion-based methods are commonly presented as targetless calibration techniques~\cite{slam, Das}. Although these targetless approaches offer greater flexibility, they still do not match the accuracy of target-based solutions~\cite{survey}. In contrast, target-based S2V or non-overlapping S2S calibration has mainly been addressed using external sensing devices~\cite{Domhof, MC, Room}. 
\begin{figure*}[!h]
    \centering
    \includegraphics[width=\textwidth]{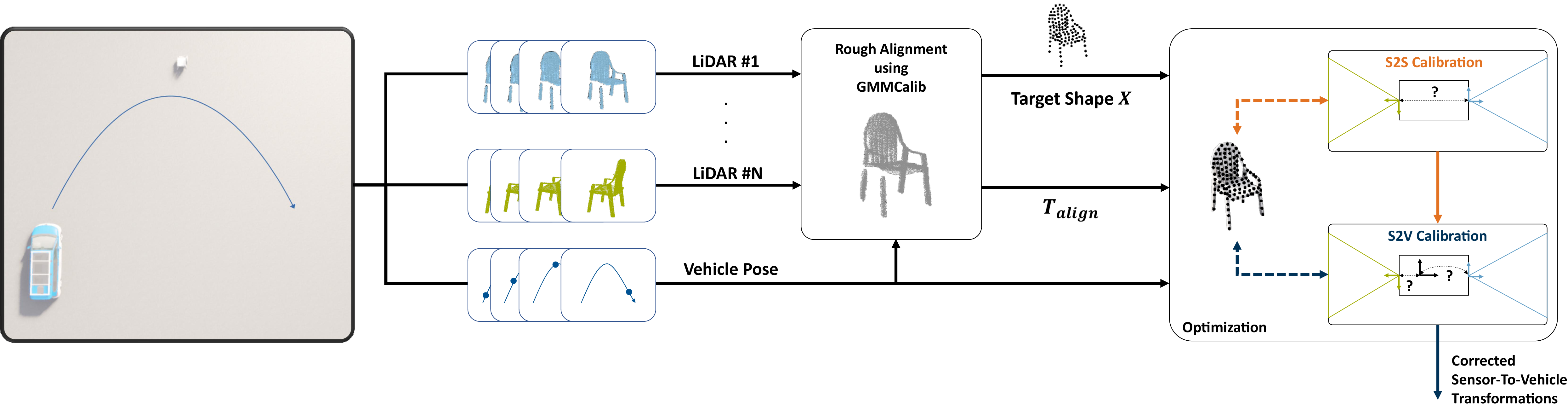}
    \caption{CaLiV: Data is collected by driving a curved trajectory. The data is first aligned using vehicle poses and different perspectives of the calibration target. The S2S calibration is first computed by the algorithm in the optimization phase, and it subsequently serves as the basis for the S2V optimization. The input to the S2S and S2V optimization is a rough point cloud alignment and the shape of the reconstructed target, both generated using GMMCalib~\cite{GMMCalib}.}
    \label{fig:Ours}
    \vspace{-13pt}
\end{figure*}

In this work, we present CaLiV, a target-based approach to address the problems of S2V calibration for an arbitrary sensor setup without any additional external sensing devices. CaLiV first performs (non-overlapping) S2S calibration and then utilizes the S2S calibration to compute the S2V calibration parameters. As the underlying alignment model, we rely on GMMCalib~\cite{GMMCalib}, which provides a specific calibration frame formulation that we utilize for simultaneous optimization of the S2V calibration parameters. The approach is illustrated in Fig.~\ref{fig:Ours}. Our contributions are as follows: 
\begin{itemize}
    \item To the best of our knowledge, we are the first to present a target-based Sensor-to-Vehicle calibration framework for a multi-LiDAR sensor setup without any external sensing devices.
    \item In addition, our approach can be used for non-overlapping Sensor-to-Sensor calibration.
    \item We use the GMMCalib formulation with a common calibration frame and generalize it to enable Sensor-to-Vehicle calibration. 
    \item We evaluate the Sensor-to-Sensor and Sensor-to-Vehicle performance of our method on a worst-case sensor setup with two LiDAR sensors facing in opposite directions in simulation and real-world experiments. We outperform the current state-of-the-art for both non-overlapping Sensor-to-Sensor and Sensor-to-Vehicle calibration.
\end{itemize}

\section{RELATED WORK}
There have been many advances in extrinsic calibration in recent years, addressing S2S and S2V calibration with different sensor setups and overlapping FoVs. To present the current state of the art, the contributions on extrinsic sensor calibration will be categorized in targetless and target-based approaches.

\subsection{Targetless Calibration}
In many cases, targetless approaches perform feature-based calibration using geometric features of the environment. For example, the works of~\cite{Wei22, Miguel23, Niijima2024}, and~\cite{lidarlink} present S2S calibration using different features such as edges, planes or corners. In the same context, feature-based approaches as presented in~\cite{Meyer2021, Kulmer2024, Heide18}, and~\cite{Yan2024} are often used to perform S2V calibration. In~\cite{Meyer2021}, road edges are used to compute the roll and pitch angle errors of the sensor relative to the robot frame. Kulmer~\etal~\cite{Kulmer2024} and Yan~\etal~\cite{Yan2024} use the ground plane to perform a partial S2V calibration that compensates for the roll and pitch angle errors. In addition, the z-coordinates are computed in~\cite{Kulmer2024} and~\cite{Yan2024} respectively.

Targetless calibration can also be performed using motion-based or mapping-based methods. The general idea of motion-based calibration is that the motion of a robot can be used to identify the extrinsics of the sensors by exploiting the rigid body assumption. This is typically done by solving a matrix equation of the form $AX = XB$, where $X$ is unknown. Solving this problem is also referred to as the hand-eye calibration problem~\cite{horaud}. Das~\etal~\cite{Das2022} use the hand-eye formulation to calibrate multiple non-overlapping LiDARs. Similarly, in~\cite{Liu2021, Chang2023}, and~\cite{Jiao19}, this formulation is used to recover the sensor extrinsics. In~\cite{Das}, a S2V framework is introduced using the hand-eye formulation to describe the relationship between the vehicle frame and the LiDAR frame.

In contrast to motion-based calibration methods, mapping-based methods use features of the environment directly for the calibration task. This can be done, for example, by simultaneously creating a map of the environment, estimating the robot's pose relative to this map, and recovering the extrinsic parameters of the sensors in a SLAM-based application as presented in~\cite{slam} and~\cite{Lin2020}. The mlcc framework presented in~\cite{Liu2022} simultaneously optimizes the LiDAR poses and sensor parameters by matching point clouds with an adaptive voxelization technique.

As mentioned in the introduction, relying on S2S calibration only is not sufficient for safe autonomous operation. In the current landscape, some of the approaches only partially fulfill these requirements by providing either S2S calibration for non-overlapping FoVs or S2V calibration. While~\cite{Das} supports S2V calibration for large vehicles using hand-eye calibration and thus implicitly calibrates non-overlapping sensor setups, the calibration does not achieve state-of-the-art translation and rotation accuracy. Although both motion-based and mapping-based calibration are well suited for non-overlapping S2S calibration, they currently do not provide accurate S2V calibration and require long motion sequences to recover the calibration parameters. S2V calibration is primarily solved by feature-based methods, which, due to the nature of the algorithms, require a feature-rich environment. 

\subsection{Target-Based Calibration}
Target-based calibration methods set up specific targets which are then scanned by all sensors. The resulting point clouds can then be matched using point cloud registration algorithms. To perform a target-based calibration, all sensors must see the target. For this reason, target-based algorithms have mostly been used to calibrate sensors with overlapping FoVs, as for example in~\cite{Zhang22, Li22}. In contrast, GMMCalib~\cite{GMMCalib} shows potential for the calibration of non-overlapping FoVs due to its robustness to different perspectives. The first target-based calibration method to perform both S2S and S2V calibration was introduced by Domhof~\etal~\cite{Domhof}. They use a Styrofoam target with circular holes for S2S calibration and an additional external laser scanner for S2V calibration. By estimating the robot's body reference frame with this external sensor, they are able to perform S2V calibration without introducing motion. To perform a target-based calibration, all sensors must be able to see the target simultaneously or motion must be introduced to create an overlapping FoV. The first option, implemented by~\cite{Room}, uses a full environment scan as a reference map, created with an external terrestrial laser scanner. By aligning individual LiDAR point clouds to this map, S2S calibration errors are resolved without moving the system. However, S2V calibration remains unaddressed since some errors require motion to be observed~\cite{Das}.

The second option is a more suitable method to also solve S2V errors. Charron~\etal~\cite{MC} used this approach to calibrate a camera-LiDAR system. Both the target and system poses are recovered using an additional motion capture system (MCS). The dimensions of the target are measured beforehand so that specific key points can be identified. The ground truth positions of these key points are known by combining the target pose returned by the MCS with the dimensions of the target. This method could theoretically solve S2V calibration errors, depending on the system’s trajectory relative to the target. Similar to~\cite{Domhof} and~\cite{Room}, the work of~\cite{MC} also presents highly accurate results with the help of an additional external sensor.

The state-of-the-art shows that for the above requirements, target-based approaches outperform targetless approaches in terms of calibration accuracy and they are not dependent on a feature-rich environment. However, this is currently only achieved with the use of external sensing devices and very specific calibration targets, which in some cases, as shown in~\cite{MC}, are even measured in advance. Accuracy is therefore achieved at the expense of the flexibility that can be provided by targetless approaches. We conclude that there is no target-based, environment-independent calibration framework for extrinsic S2V calibration of multi-LiDAR systems with non-overlapping FoVs. Therefore, we propose a target-based calibration framework that enables both S2V and non-overlapping calibration. We achieve this without the use of additional external sensor systems.

\section{METHODOLOGY} \label{sec:methodology}
This section describes the methodology behind the framework of Fig.~\ref{fig:Ours}. For the sake of clarity, we consider the calibration with a vehicle with two non-overlapping LiDARs. However, the proposed technique can easily be generalized for any multi-LiDAR setup.
\subsection{Spatial Relationships}\label{sec:problem_statement}
\begin{figure}[!h]
\includegraphics[width=\linewidth]{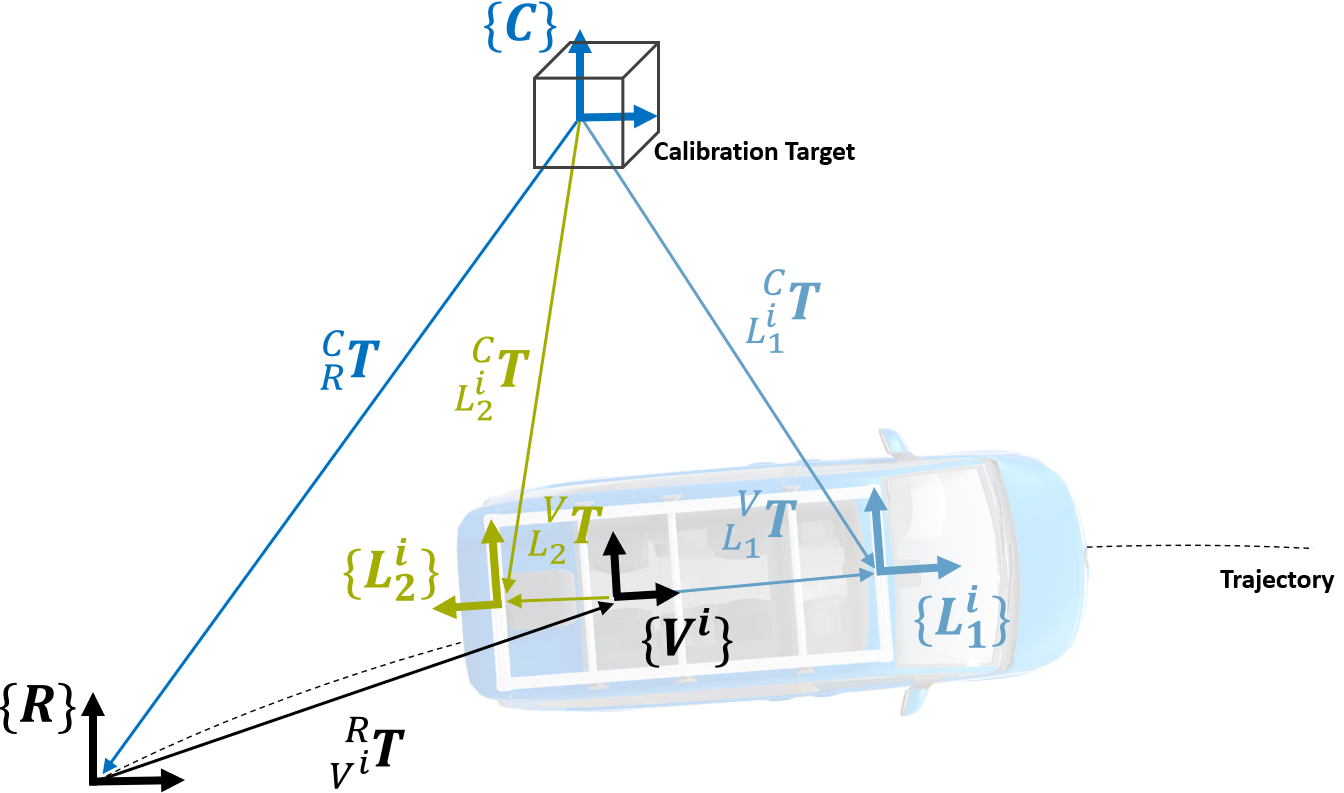}
\centering
\caption{Spatial relationship between the frames $\{V^i\}$ of a moving system at a specific time steps $t_i$ with two LiDAR frames $\{L^i_1\}$ and $\{L^i_2\}$, a global reference frame $\{R\}$, and a common calibration frame $\{C\}$.}
\label{fig:spat_rel}
\vspace{-0.5cm}
\end{figure}
Fig. \ref{fig:spat_rel} shows the spatial relationships, relevant for the calibration task, for a moving vehicle $V$ with two non-overlapping LiDARs $L_1$ and $L_2$. $\{L_1^i\}$, $\{L_2^i\}$ and $\{V^i\}$ describe the respective LiDAR and vehicle frames at a time step $t_i$ with $1 \leq i \leq N$. $\{R\}$ represents the global reference frame and $\{C\}$ the common calibration frame as introduced in~\cite{GMMCalib}. 

${}_{L_1}^{V}T$ and ${}_{L_2}^{V}T$ describe the ground-truth transformations of the LiDARs relative to $\{V^{i}\}$ at all time steps. Recovering these ground-truth values is the goal in our calibration. Initially, we are only given erroneous LiDAR transformations, which will be referred to as ${}_{L_1}^{V}\hat{T}$ and ${}_{L_2}^{V}\hat{T}$ in the following. Further, ${}_{V^i}^{R}T$ represents the vehicle pose at time step $t_i$ relative to the global reference frame and must be known before the calibration. 

In addition to the relationships depicted in Fig. \ref{fig:spat_rel}, we choose $L_2$ to be the reference sensor in our setup. Therefore, we define the transformation $^{L_2}_{L_1}T$ to be the ground truth S2S transformation and $^{L_2}_{L_1}\hat{T}$ the initial S2S transformation. The S2V transformation of $L_1$ can then be defined as: 

\begin{equation}\label{eq:S2Srel}
^{V}_{L_1}T = ^{V}_{L_2}T~^{L_2}_{L_1}T
\end{equation}

\subsection{Rough Point Cloud Alignment}\label{sec:alignment}
As introduced in GMMCalib~\cite{GMMCalib}, the point clouds are not aligned by matching source and target point clouds but rather by registering point clouds of different frames to a latent calibration frame. Accordingly, ${}_{L^i_1}^{C}T$ and ${}_{L^i_2}^{C}T$ are transformations that align the point clouds of a LiDAR at a specific time step in the common calibration frame $\{C\}$. 

Before applying GMMCalib, we perform a pre-alignment step by transforming the point clouds with the initial LiDAR transformations ${}_{L_1}^{V}\hat{T}$ and ${}_{L_2}^{V}\hat{T}$, respectively, as well as the ego vehicle pose ${}_{V^i}^{R}T$ at the measurement's time step. In total, the alignment transformations for the LiDARs are defined by the following equation:
\begin{equation}
\begin{split}
{}_{L^i_a}^{C}T = {}_{R}^{C}\hat{T} ~ {}_{V^i}^{R}T ~ {}_{L_a}^{V}\hat{T} \quad \quad \quad \\\forall i, a \quad s.t. \quad 1 \leq i \leq N, ~ a \in \{1,2\}.
\end{split}
\end{equation}
${}_{R}^{C}\hat{T}$ here is the part of the alignment that is done by the point cloud registration algorithm of GMMCalib and $a$ represents the specific LiDAR sensor. Similar to the ICP algorithm, the registration of GMMCalib requires a good initial alignment of the point clouds as it positively affects the calibration results. This is achieved by the pre-alignment step discussed above.
\subsection{Non-Linear Optimization}\label{sec:opt}
${}^{C}_{R}T$ represents the pose of the reference frame $\{R\}$ relative to the calibration frame $\{C\}$. Note that this latent frame formulation allows us to compute the Sensor-to-Vehicle calibration in the first place by assuming a fixed transformation between the calibration frame and the reference frame. In this specialized dual-LiDAR version of the calibration problem, ${}^{C}_{R}T$ can be computed as follows:
\begin{equation}
\begin{split}
{}_{R}^{C}T = {}_{L_{a}^{i}}^{C}T ~ {}_{L_a}^{V}T^{-1} ~ {}_{V^{i}}^{R}T^{-1} \quad\quad\\
\forall i, a \quad s.t. \quad 1 \leq i \leq N, ~ a \in \{1,2\}.
\end{split}
\end{equation}
This simply follows by the relationships depicted in Fig. \ref{fig:spat_rel}.

Since the LiDARs' ground truth S2V transformation ${}_{L_2}^{V}T$ and S2S transformation ${}_{L_1}^{L_2}T$ are unknown, we cannot compute ${}_{R}^{C}T$ directly. Therefore, we introduce ${}^C_R\hat{T}_a^i$ which describes the approximation of ${}^C_RT$ for the LiDAR $L_a$ at time step $t_i$ for specific estimates ${}_{L_2}^{V}T^{'}$ and ${}_{L_1}^{L_2}T^{'}$ of the LiDAR transformations. Note that these estimates can differ from the initial transformations and are chosen in an iterative minimization process. Then, we can define an error function $e({}_{R}^{C}\hat{T}^i_a, {}_{R}^{C}\hat{T}^j_b)$ which returns the error between a pair of approximated transformations between reference and calibration frame.

In a non-overlapping setup, the respective LiDARs may not see the target at all time steps and thus we cannot simply sum the results of the error function for all LiDARs at all time steps to evaluate our current LiDAR transformation estimates. Therefore, we define a set $S$ as the set of all pairs of sensors and time steps for which such a transformation is given, i. e. the sensor sees the target. Specifically, we say $(i, a) \in S$ \textit{iff} ${}_{L^i_a}^{C}T$ is known.

Recovering the optimal transformations ${}_{L_2}^{V}T^\star$ and ${}_{L_1}^{L_2}T^\star$ can now be reduced to the following minimization problem:

\begin{equation}\label{eq:errorfunction}
{}_{L_2}^{V}T^\star, {}_{L_1}^{L_2}T^\star = \operatorname*{arg\,min}_{{}_{L_2}^{V}T^{'}, {}_{L_1}^{L_2}T^{'}}
\ \sum_{\{(i, a), ~ (j, b)\} \subset S} 
    e\!\left({}_{R}^{C}\hat{T}^{i}_{a}, ~ {}_{R}^{C}\hat{T}^{j}_{b}\right).
\end{equation}

This same minimization problem can then be formulated with a fixed S2S transformation ${}_{L_1}^{L_2}T^\star$ in a second iteration, which further improved S2V calibration results in our experiments.

\subsection{Implementation Details}
We now present the implementation details of our framework used to solve the minimization problem formulated in Section \ref{sec:opt}. Specifically, we discuss the kinematic model used to recover the vehicle poses ${}_{V^{i}}^{R}T$, the joint registration algorithm used to obtain the transformations ${}_{L^i_a}^{C}T$ of all point clouds to a common calibration frame $\{C\}$, as well as the minimization algorithm.
\subsubsection{Kinematic Model}
The approach is compatible with any kinematic model, provided that the vehicle poses are given at the time steps of the LiDAR measurements. We use an Unscented Kalman Filter (UKF) for the kinematic model that combines IMU, GPS, velocity and orientation data to recover vehicle pose estimates ${}_{V^{i}}^{R}T$ at all time steps $t_i$. We assume these sensors to be calibrated. Our UKF also allows the modeling of sensor-specific noise and uncertainties to evaluate the calibration accuracy under noisy sensor data. 
\subsubsection{Joint Registration}
Before a point cloud registration algorithm can be applied to the point clouds, they must be pre-processed. First, the point clouds are transformed with the initial LiDAR transformations and the ego vehicle pose at the measurement's time step, as discussed in Section \ref{sec:alignment}. Then, we separate the target points from the ground points by using the Random Sample Consensus (RANSAC)~\cite{ransac} algorithm to identify and filter out the ground plane. Finally, we use GMMCalib~\cite{GMMCalib} for joint registration of the filtered point clouds into a common calibration frame $\{C\}$. GMMCalib returns the registration transform ${}_{L^i_1}^{C}T$ and ${}_{L^i_2}^{C}T$ for each point cloud that contains points belonging to the target shape.
\subsubsection{Minimization Algorithm}
As shown in Eq. \ref{eq:errorfunction}, the cost function is formulated as the sum of all pairwise errors of the estimated transformations between calibration frame and reference frame. Note that the previously discussed problem statement assumes that the vehicle poses and calibration transformations are accurate. To account for errors introduced by the registration algorithm or the kinematic model, we introduce an outlier filter that removes the most extreme outliers of the computed errors. For this purpose, we introduce the parameter $k$. After computing all pairwise errors between estimates of ${}_{R}^{C}T$, we disregard all errors that are below the $k^{th}$ percentile or above the $(100-k)^{th}$ percentile. We set $k$ to 10 in our implementation, as this proved to produce the best results for our experiments.

The application of an outlier filter introduces a non-differentiable cost function, as increasing or decreasing an error value may cause discontinuous transitions in and out of the defined percentiles. Therefore, a minimization algorithm that can handle such functions is needed. Consequently, Powell's conjugate direction method~\cite{Powell} is implemented to solve the minimization problem.

\begin{figure}
    \centering
    \includegraphics[width=1\linewidth]{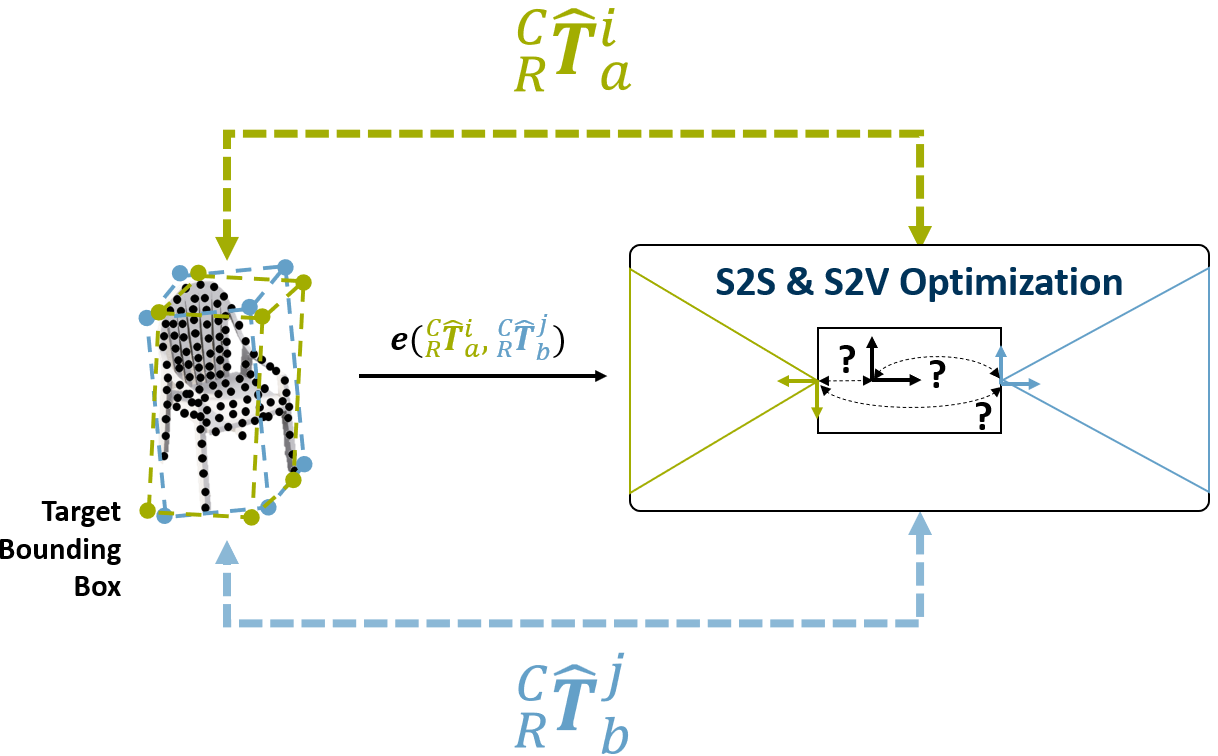}
    \caption{This figure illustrates the optimization process of the calibration. The two transformation estimates are used to transform the point sets represented by the bounding box of the calibration target shape.}
    \label{fig:error_viz}
    \vspace{-0.5cm}
\end{figure}

The error function $e$ introduced in Section~\ref{sec:opt} computes the error between two estimates ${}_{R}^{C}\hat{T}^{i}_{a}$ and ${}_{R}^{C}\hat{T}^{j}_{b}$ of the transformation between calibration frame and reference frame. We invert the transformations to be able to apply them to a set of points in the calibration frame $\{C\}$. To calculate the error between two transformations, we transform the set of points and compute the their squared pairwise distances. With GMMCalib returning the reconstructed shape of the target object, we define the points of the oriented bounding box of this shape to be the point set for this error function as illustrated in Figure~\ref{fig:error_viz}. In contrast to the complete reconstructed target point cloud, the bounding box representation of the point set decreases the performance overhead in non-linear optimization.

\section{EXPERIMENT DESIGN}\label{sec:experiments}
In this chapter, we present the experimental setup used to evaluate our proposed framework. The simulation is performed in the driving simulator CARLA~\cite{CARLA} and consists of the digital twin of the real-world vehicle EDGAR~\cite{edgar}, a generic calibration target, and a flat surface. To validate our approach, the experiments are also performed in a real-world environment. Furthermore, we describe the evaluation metrics of the S2S and S2V calibration performance.
\subsection{Simulation \& Real World Setup }\label{section:sim_setup}
In both the simulation and the real world, the vehicle is equipped with two LiDARs with a non-overlapping FoV. One LiDAR is mounted on the front end of the vehicle roof at the position \(\begin{bmatrix} 2 & 0 & 2 \end{bmatrix}^T\) and is rotated $-10\degree$ around the y-axis. It, therefore, faces the driving direction and is tilted to the ground. The other LiDAR is mounted on the roof's rear end at the position \(\begin{bmatrix} -0.4 & 0 & 2 \end{bmatrix}^T\) and is also rotated $-10\degree$ around the y-axis and additionally $180\degree$ around the z-axis. This means that the rear LiDAR is facing the opposite direction of travel and is also tilted towards the ground. These exact mounting positions are given for the simulation, but correspond to the expected positions of the real vehicle. The characteristics of the simulated sensors, listed in Table \ref{tab:sensor_characteristics_noises}, are chosen to match those of real-world sensors. The resulting LiDAR FoVs are qualitatively shown in the optimization part of Fig.~\ref{fig:Ours}.
\begin{table}[h]
    \centering
    \small 
    \resizebox{0.45\textwidth}{!}{ 
    \begin{tabular}{@{}clcc@{}} 
    \toprule
    \textbf{Sensor} & \textbf{Attribute}  & \textbf{Dimension} &\textbf{Value} \\ 
    \midrule
    LiDAR & Horizontal FoV [°] & - & 120 \\ 
    LiDAR & Vertical FoV [°] & - & 25 \\ 
    LiDAR & Range [m] & - & 500 \\ 
    LiDAR & Channels [-] & - & 152 \\ 
    LiDAR & Noise [$m$] &  (x,y,z) & 0.01 \\ 
    Accelerometer & Noise [$\frac{m}{s^{2}}$] & (x,y,z)   & 0.05 \\ 
    Gyroscope & Noise [$\frac{rad}{s}$]    & (x,y,z) & 0.005 \\ 
    GPS & Noise [$\degree$] & (lat,lon/alt) & 1.5e-7/0.02 \\ 
    Orientation & Noise [$rad$]  & (x,y,z) & 0.005 \\ 
    Velocity & Noise [$\frac{m}{s}$]  &  (x,y,z) & 0.1 \\ 
    \bottomrule
    \end{tabular}
    }
    \caption{Sensor Characteristics of the CARLA simulation}
    \label{tab:sensor_characteristics_noises}
    \vspace{-6pt}
\end{table}

\begin{figure}[h] 
    \includegraphics[scale=0.23]{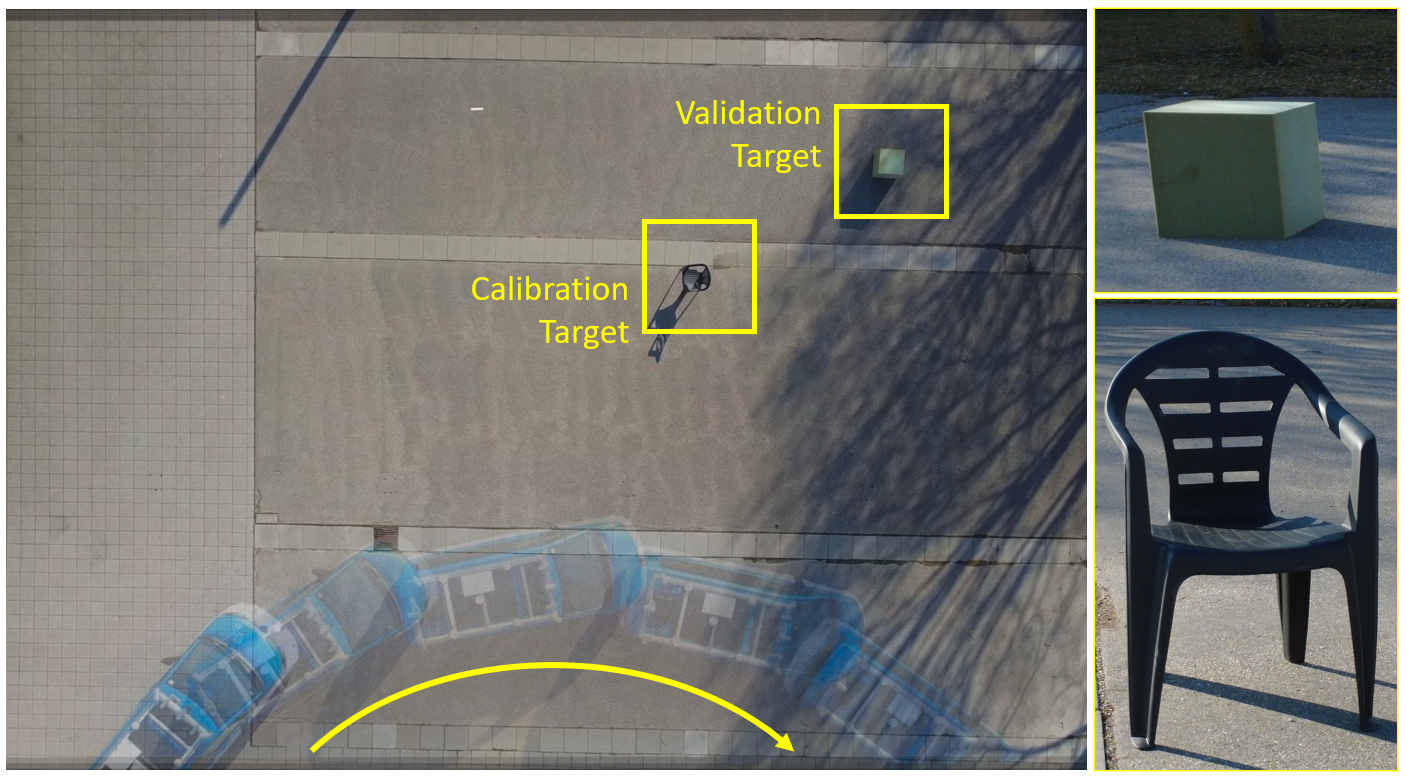}
    \centering
    \caption{\textbf{Left}: Curved maneuver with the test vehicle in the real-world setup. \textbf{Right}: A chair as a calibration target and the cubic validation target.}
    \label{fig:RealEnv}
    \vspace{-8pt}
\end{figure}

The simulated vehicle is also equipped with an IMU (accelerometer and gyroscope) and GPS. All sensors and additional data are augmented by an artificially added Gaussian noise. As a calibration target we use a plastic chair. It should be noted, however, that any target can be used for these experiments. The only requirements are that the target should be detectable by the LiDAR sensor and, for non-overlapping FoVs, it is a prerequisite to use non-symmetric targets with respect to the observed perspectives. In the real-world experiments, an additional cubic target is placed at a position that is not part of the optimization as described in Section~\ref{sec:methodology} to serve as a validation target (see Fig.~\ref{fig:RealEnv}).
Another important consideration is the robot maneuver, since certain S2V errors are not observable for some trajectories. We use the work of \cite{Das} as a basis for these considerations, which primarily uses curves for S2V calibration. The trajectory used in the simulation is shown in Fig.~\ref{fig:Ours} and was executed with an average velocity of $2\frac{m}{s}$ and a static steering angle of about $17\degree$. These values are chosen such that they fulfill maximum observability for the S2V calibration and require minimal space for the robot maneuver. The real-world setup is shown in Fig.~\ref{fig:RealEnv} and is analogous to the simulated environment.

\subsection{Evaluation}
In the simulation, the ground truth LiDAR transformations relative to the vehicle frame are known and denoted as ${}^V_{L_1}T$ and ${}^V_{L_2}T$ for LiDAR $L_1$ and $L_2$, respectively. After the optimization, we recovered the optimized versions ${}_{L_2}^{V}T^\star$ of the $L_2$ S2V transformation and ${}_{L_1}^{L_2}T^\star$ of the S2S transformation. To obtain the optimized $L_1$ S2V transformation ${}_{L_1}^{V}T^\star$, Eq. \ref{eq:S2Srel} can be used. The S2V errors left after the optimization step can now simply be defined as the residuals of ${}_{L_1}^{V}T^\star$ and ${}_{L_2}^{V}T^\star$ relative to ${}^V_{L_1}T$ and ${}^V_{L_2}T$, respectively.

For the S2S evaluation, we again define $L_2$ as the reference sensor and, therefore, define its S2V transformation ${}^{V}_{L_2}T^*$ as fixed. This optimized transformation still contains a S2V error. The following expression describes the transformation of $L_1$ relative to the vehicle frame which contains no S2S error but still contains the S2V error of ${}^{V}_{L_2}T^*$:
\begin{equation}
    {}^{V}_{L_2}T^* ~ {}^{L_2}_{L_1}T
\end{equation}
The S2S calibration error can now be defined as the residual of the optimized ${}^{V}_{L_1}T^*$ (which still contains the S2S error) relative to the expression above, since both share the same S2V error. This way, the S2S error can be evaluated independently from the S2V error. In our experiments, we added randomly generated errors to the ground truth transformations of $L_1$ and $L_2$ to obtain different initial errors on the calibration matrices. The translation errors for x, y and z are uniformly distributed in the range of $-0.1m$ to $0.1m$. The rotation errors for roll, pitch and yaw are also uniformly distributed, but in the range of $-3.0\degree$ to $3.0\degree$. In total, we run 100 simulations containing noisy UKF poses.
\section{RESULTS \& DISCUSSION}

\subsection{S2S Calibration}
In this section, we evaluate our method regarding S2S calibration and compare our method with the targetless open-source algorithm mlcc~\cite{Liu2022}. Due to the fact that no comparable target-based approach currently exists, we compare this targetless framework because it supports non-overlapping FoVs. For the sake of comparison, we will perform the motion proposed in Section~\ref{section:sim_setup} for both algorithms. The work of~\cite{Liu2022} is evaluated with ground truth vehicle poses. The mean calibration errors of all executions are shown in Table~\ref{tab:sts_curved}.

\begin{table}[h]
    \centering
    \setlength{\tabcolsep}{5pt} 
    \begin{tabular}{lcccccc}
    \toprule
    \textbf{Algorithm} & $\delta x$ & $\delta y$ & $\delta z$ & $\delta \phi$ & $\delta \theta$ & $\delta \psi$ \\
    \midrule
    mlcc~\cite{Liu2022} & 5.5642 & 0.4691 & 2.2511 & 0.1116 & 0.3222 & 0.1286 \\
    CaLiV & \textbf{0.0011} & \textbf{0.0154} & \textbf{0.0200} & \textbf{0.0043} & \textbf{0.0070} & \textbf{0.0005} \\
    \bottomrule
    \end{tabular}
    \caption{Mean S2S translation and rotation errors in $m$ and $rad$, respectively, using UKF poses. mlcc comparison is performed using ground truth poses.}
    \label{tab:sts_curved}
    \vspace{-6pt}
\end{table}

The mlcc framework~\cite{Liu2022} is an accurate method for non-overlapping S2S calibration. However, in our experimental setup, it yielded significantly less accurate results, with mean translation errors reaching several meters and mean rotation errors ranging from $6^\circ$ to $19^\circ$. While mlcc works very well when given accurate initial extrinsics and poses, it does not provide accurate results when exposed to erroneous LiDAR poses that are due to S2V calibration errors. In their work, Liu~\etal~state that the initial LiDAR extrinsic parameters should be pre-calibrated and accurate LiDAR poses must be available, e.g. by first applying a hand-eye calibration or SLAM. From our results, we conclude that the initial S2V errors of up to $3\degree$ introduced in our experiment cannot be handled by mlcc. Therefore, in contrast to CaLiV, a more accurate prior calibration is necessary.

\begin{figure}[t]
\centering
    \begin{subfigure}{0.23\textwidth}
        \centering
        \includegraphics[width=\linewidth]{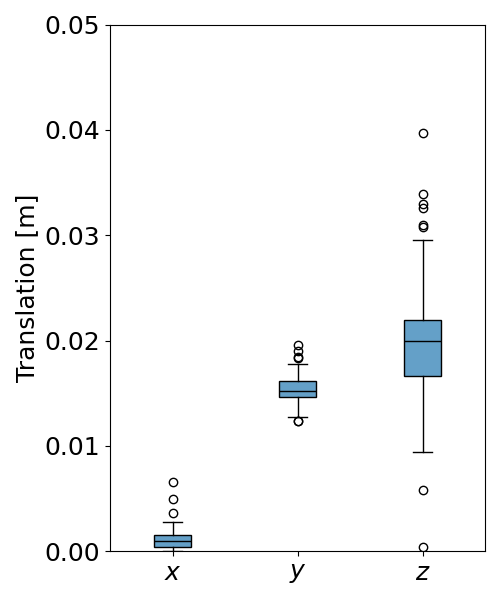}
    \end{subfigure}
    \begin{subfigure}{0.24\textwidth}
        \centering
        \includegraphics[width=\linewidth]{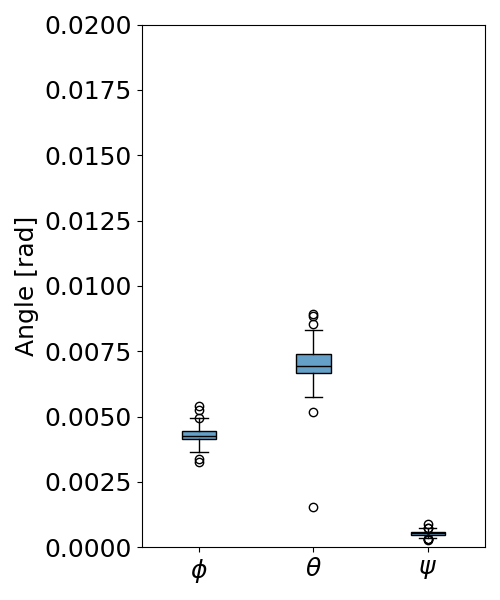}
    \end{subfigure}
    \caption{The distribution of the non-overlapping S2S translation and rotation errors of CaLiV.}\label{fig:s2s_res}
    \vspace{-12pt}
\end{figure}

Fig.~\ref{fig:s2s_res} shows the calibration errors' distributions of our proposed algorithm. Although there are outliers, they are relatively close to the median. Furthermore, the standard deviations are relatively low for all errors. The maximum error of all calibrations is below $0.6\degree$ and the median is around $0.43\degree$.

An important question when calculating the S2S calibration is why it is even needed, since the S2V calibration is performed for both sensors. Experimental results show that performing S2S calibration as a first step improves the overall accuracy of the subsequent S2V calibration. This pre-alignment improves the consistency between sensor measurements, resulting in more accurate S2V calibration results. To demonstrate the robustness of our approach against extreme differences in perspective, we chose a sensor setup representing the worst-case scenario of non-overlapping sensor coverage, with two sensors facing opposite directions. 

\subsection{S2V Calibration}
Correcting the S2V transformation errors depends heavily on the vehicle's trajectory and the distance to the calibration target. With our proposed trajectory, we found that the S2V translation parameters could not be reliably estimated due to the parameters' lack of observability. Therefore, CaLiV is designed to estimate only the rotation parameters. This topic will be discussed further in Section~\ref{sec:observability}.
\begin{figure}[!h]
\centering
    \begin{subfigure}{0.23\textwidth}
        \centering
        \includegraphics[width=\linewidth]{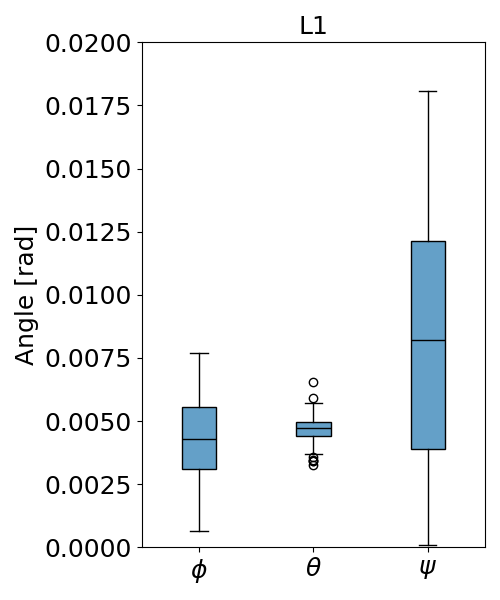}
    \end{subfigure}
    \begin{subfigure}{0.23\textwidth}
        \centering
        \includegraphics[width=\linewidth]{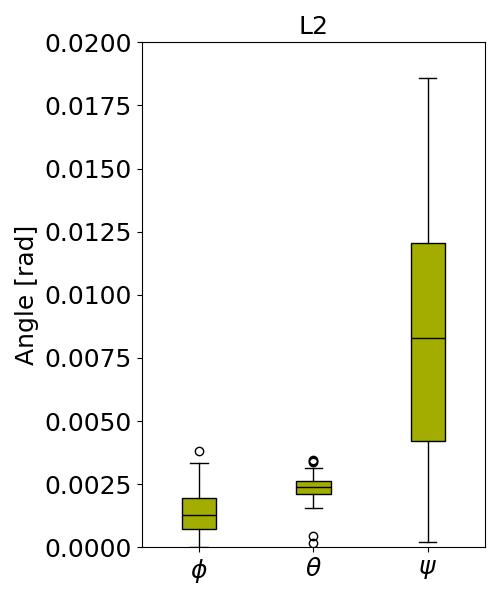}
    \end{subfigure}

    \caption{The distribution of the S2V rotation errors of CaLiV. The distributions for $L_1$ are shown on the left, while the right plot shows the distributions of the $L_2$ errors.}
    \label{fig:s2v_res}
    \vspace{-5pt}
\end{figure}

In the following, we present the S2V calibration results and distinguish between the $L_1$ and $L_2$ S2V calibration error. Since $L_2$ is the reference LiDAR, the $L_1$ S2V error is dependent on the $L_2$ S2V error and the S2S error. As described in Section~\ref{sec:methodology}, the second stage of the calibration process makes use of the previously calibrated S2S transformations. For the S2V calibration, an additional constraint must be introduced: To account for the observability issues and to compute the rotation errors independently, the S2V translation errors are fixed to zero. 

Table \ref{tab:stv_curve2} shows the mean $L_1$ and $L_2$ S2V roll ($\phi$), pitch ($\theta$) and yaw ($\psi$) errors after performing 100 calibrations. The roll and pitch errors are both less than $0.1 \degree$. The mean S2V yaw errors of $L_1$ and $L_2$ are all below $0.5\degree$. Fig. \ref{fig:s2v_res} shows the distributions of 100 $L_1$ (left figure) and $L_2$ (right figure) S2V rotation errors in radian after the S2V calibration. The roll and pitch distributions have a low standard deviation and few outliers. All roll and pitch errors are below $0.5\degree$. Although the standard deviation of the yaw errors and the number of outliers are noticeably bigger, the mean is close to the median and the maximum errors are a little over $1\degree$.
\begin{table}[!h]
    \centering
    \begin{tabular}{lcccc}
    \toprule
    \textbf{LiDAR}  & \textbf{$\delta \phi$} & \textbf{$\delta \theta$} & \textbf{$\delta \psi$}\\
    \midrule
    \multirow{1}{*}{$L_1$} & 0.0042 & 0.0047 & 0.0082 \\ \\
    \multirow{1}{*}{$L_2$} & 0.0014 & 0.0024 & 0.0083 \\ 
    \bottomrule
    \end{tabular}
    \caption{Mean S2V rotation errors in $rad$ of this work's algorithm for the curved trajectory. The proposed method is evaluated for UKF poses.}
    \label{tab:stv_curve2}
        \vspace{-5pt}
\end{table}

The S2V calibration performance of CaLiV is compared to the two open-source calibration methods Multi-LiCa~\cite{Kulmer2024} and SensorX2Car~\cite{Yan2024}. Both methods compute the partial LiDAR-to-Vehicle calibration using the ground plane to estimate the pitch and roll angles. The results are shown in Table~\ref{tab:benchmark}. Note that Multi-LiCa and SensorX2Car do not provide the calibration transformation for all rotation angles, and therefore, only the roll estimates for Multi-LiCa and the roll and pitch angles for the SensorX2Car calibration algorithm can be compared. For the comparison with CaLiV, we use the ground truth poses to evaluate the calibration accuracy only.
\begin{table}[h]
    \centering
    \begin{tabular}{lcccccc}
    \toprule
    \textbf{Algorithm} & $\delta \phi$ & $\delta \theta$ & $\delta \psi$ \\
    \midrule
    Multi-LiCa~\cite{Kulmer2024} & - & 0.0056 & -\\
    SensorX2Car~\cite{Yan2024} & 0.0046 & 0.0019 & - \\
    CaLiV  & \textbf{0.0013} & \textbf{0.0007} & \textbf{0.0084}\\
    \bottomrule
    \end{tabular}
    \caption{Mean rotation errors in $rad$ for the curved trajectory. To compare the calibration performance only the calibration was performed using ground truth poses.}
    \label{tab:benchmark}
    \vspace{-7pt}
\end{table}
\subsection{Observability} \label{sec:observability}
We want to highlight the accuracy of the calibration with respect to the rotational components, as well as discuss the translational components. In our experiments, we found that the z-translation component is not observable with a planar trajectory. However, the z-translation can easily be compensated for using ground plane information, as presented in~\cite{Kulmer2024}. Another approach to handling the z-translation was proposed by iCalib~\cite{iCalib}, which used a physical ramp to introduce the necessary excitation in that dimension. 

More notably, due to the nature of the trajectory, the algorithm cannot distinguish between translational $x$-error and yaw error. In experiments, where the translation errors were not constrained to zero, we observed a calibration error for the reference sensor of $\delta x = 0.2792~m$ and $\delta \phi=0.0304~rad$. With constrained translation errors the yaw angle can reliably be estimated with the accuracy reported in Table~\ref{tab:stv_curve2}. This could also be solved by driving longer trajectories, as demonstrated in~\cite{Das}. However, the target-based approach limits the length and topology of the trajectory. This creates a trade-off in which the method prioritizes compensating for rotational calibration errors. It should be noted that the error of the $y$-component of the translation vector is not subject to the observability issues and achieves a mean accuracy of $0.0160~m$.

The rotational errors are considered to be the most important for autonomous driving tasks, since they increase with distance and therefore pose a safety concern~\cite{GMMCalib}. More specifically, from a purely geometric point of view, a S2V yaw error of $1\degree$ corresponds to a lateral displacement of an object of nearly $2m$ at a distance of $100m$. Since translation errors remain constant over distance, we assume that there are no translation errors and focus on the rotation errors in the S2V optimization. At the cost of possible S2V translation errors, this assumption leads to more accurate S2V calibration results overall.

\subsection{Real World}
We validate CaLiV in real-world experiments as described in Section~\ref{sec:experiments}. Since no ground truth of the transformation between the front and rear LiDAR is available in real-world environments, the real-world performance is evaluated qualitatively on the validation target. 
\begin{figure}[h] 
    \includegraphics[scale=0.243]{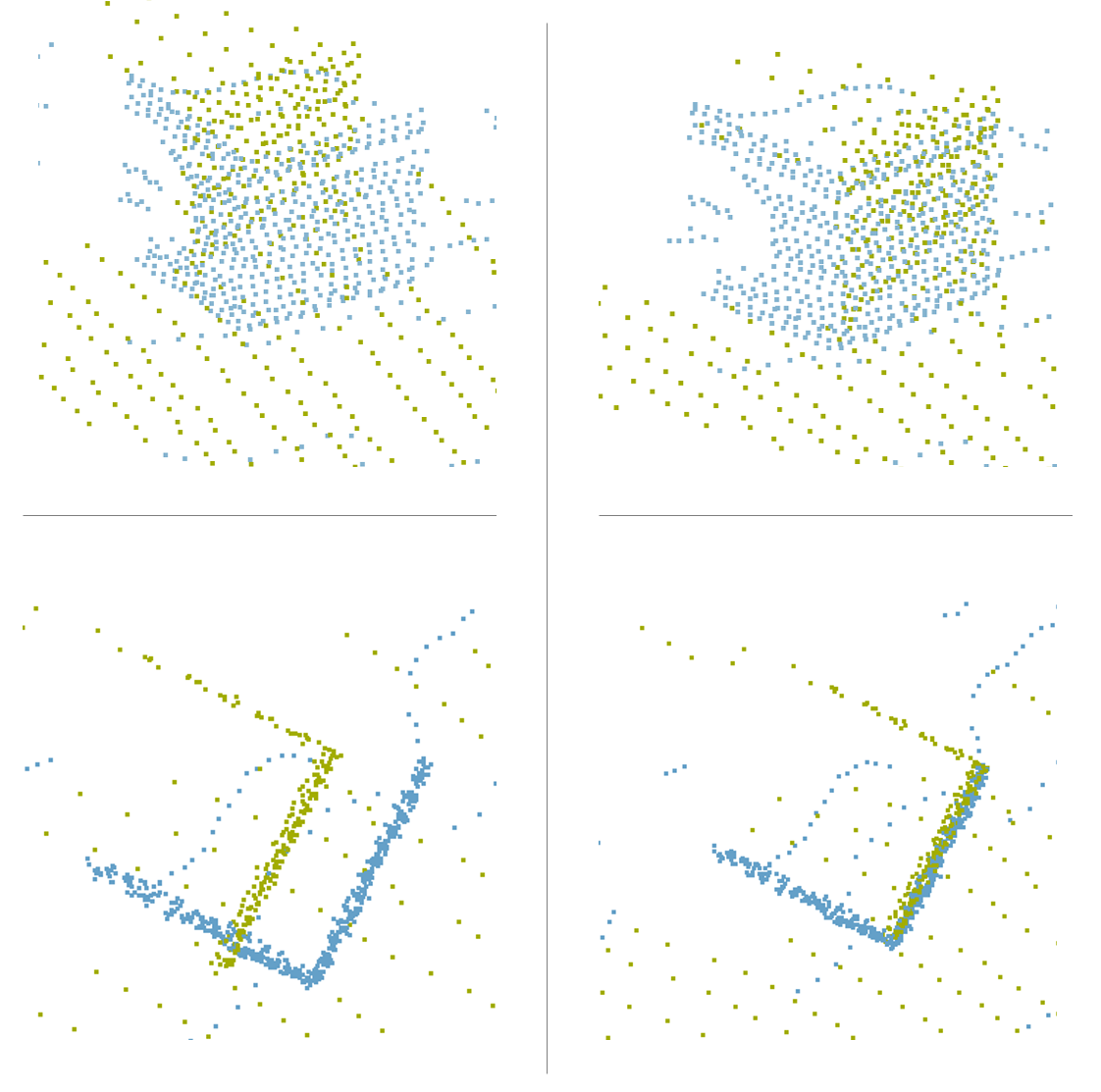}
    \centering
    \caption{\textbf{Left}: Point cloud of the validation target before the calibration (green: rear LiDAR, blue: front LiDAR). \textbf{Right}: Point clouds of the validation target after the calibration.}
    \label{fig:RealRes}
    \vspace{-8pt}
\end{figure}
It should be noted that no prior sensor calibration was performed, resulting in relatively high initial calibration errors as shown on the left of Fig.~\ref{fig:RealRes}. The figure shows the first frame of the front LiDAR at the beginning of the trajectory and the last frame of the rear LiDAR at the end of the trajectory. This gives a better insight into the overall ability of CaLiV to compensate for the propagating S2V calibration error over time/distance. After performing the calibration with CaLiV, the validation target seen on the right side of Fig.~\ref{fig:RealRes} is well aligned, indicating a successful S2V calibration. To achieve these results, a number of criteria must be met in the real experiments. To account for the motion distortion of the LiDAR sensors and the distortion of the angular velocities during the curved trajectory, the vehicle motion must be stationary and at a constant velocity of about 1.4 $\frac{m}{s}$. At this velocity, a 20~Hz LiDAR sensor, and  with the object covering only a small fraction of the LiDAR FoV, the effect of motion distortion can be mitigated for our experiments. Furthermore, real-world LiDAR sensors do not have homogeneous point resolution across the vertical and horizontal FoV, resulting in sparse point clouds at the edges of the FoV. In our experiments, this required an additional voxel and outlier filter to account for the differences in cardinality and improve the registration results during calibration.   
\section{CONCLUSION \& FUTURE WORK}
The goal of this work was to provide a target-based framework for S2S and S2V calibration of multi-LiDAR systems with arbitrary FoV overlap. In addition, we aimed to develop an algorithm that is independent of environmental features and does not rely on external sensors. CaLiV demonstrates high accuracy and robustness for both S2S and S2V calibration. CaLiV outperforms the current state of the art in both non-overlapping sensor calibration and S2V calibration. Our current method requires a curved trajectory to optimize S2S and S2V calibration. For the S2S errors it is also possible to move the target instead of the vehicle and it remains to be investigated whether this would yield equally accurate results. Furthermore, the possibility of using a more complex trajectory could be explored to also solve the S2V translation errors while maintaining the accuracy.


\bibliography{SensorToVehicle.bib}
\bibliographystyle{IEEEtran}

\end{document}